\documentclass[10pt,twocolumn,letterpaper]{article}

\usepackage{iccv}
\usepackage{times}
\usepackage{epsfig}
\usepackage{graphicx}
\usepackage{amsmath}
\usepackage{amssymb}

\usepackage[pagebackref=true,breaklinks=true,letterpaper=true,colorlinks,bookmarks=false]{hyperref}

\iccvfinalcopy 

\def\iccvPaperID{****} 

\ificcvfinal\fi

\let\ACMmaketitle=\maketitle
\renewcommand{\maketitle}{\begingroup\let\footnote=\thanks \ACMmaketitle\endgroup}

\begin{document}

\title{Learning Visual Relation Priors for Image-Text Matching and Image Captioning with Neural Scene Graph Generators}

\author{Kuang-Huei Lee$^*$$\dag$\let\thefootnote\relax\footnote{Kuang-Huei is now at Google Brain. Email: leekh@google.com} \qquad Hamid Palangi$\dag$\let\thefootnote\relax\footnote{Equal contributions. The names are shown in alphabetical order.} \qquad Xi Chen \qquad Houdong Hu \qquad Jianfeng Gao\\
Microsoft AI and Research\\
{\tt\small \{kualee,hpalangi,chnxi,houhu,jfgao\}@microsoft.com}
}

\maketitle


\begin{abstract}
Grounding language to visual relations is critical to various language-and-vision applications.
In this work, we tackle two fundamental language-and-vision tasks: \textnormal{image-text matching} and \textnormal{image captioning}, and demonstrate that neural scene graph generators can learn effective visual relation features to facilitate grounding language to visual relations and subsequently improve the two end applications.
By combining relation features with the state-of-the-art models, our experiments show significant improvement on the standard Flickr30K and MSCOCO benchmarks.
Our experimental results and analysis show that relation features improve downstream models' capability of capturing visual relations in end vision-and-language applications.
We also demonstrate the importance of learning scene graph generators with visually relevant relations to the effectiveness of relation features.
\end{abstract}


\section{Introduction}
\label{sec:intro}

Vision-and-language refers to a range of tasks that bridge vision and natural language, e.g. automatically describing visual content with text.
Early neural approaches to vision-and-language \cite{kiros2014unifying,vinyals2015show,nam2016dual,xu2015show,faghri2017vse++} often encode visual information with pre-trained classification networks like VGG-Net \cite{simonyan2014very} and ResNet \cite{he2016deep}.
Recently, Anderson \textit{et al.} \cite{anderson2017bottom} demonstrated that image understanding at instance-level can provide valuable prior knowledge to help language-and-vision models focus on salient objects and stuffs.
This ``bottom-up attention`` approach (as the attention on salient objects and stuffs comes bottom-up from perceptional priors instead of textual context) has proven to be very successful across various tasks including visual question answering, caption generation, image-text matching, and text-to-image synthesis \cite{anderson2017bottom,lee2018stacked,li2019object}.

\begin{figure}[t]
\begin{center}
\includegraphics[width=1.0\linewidth]{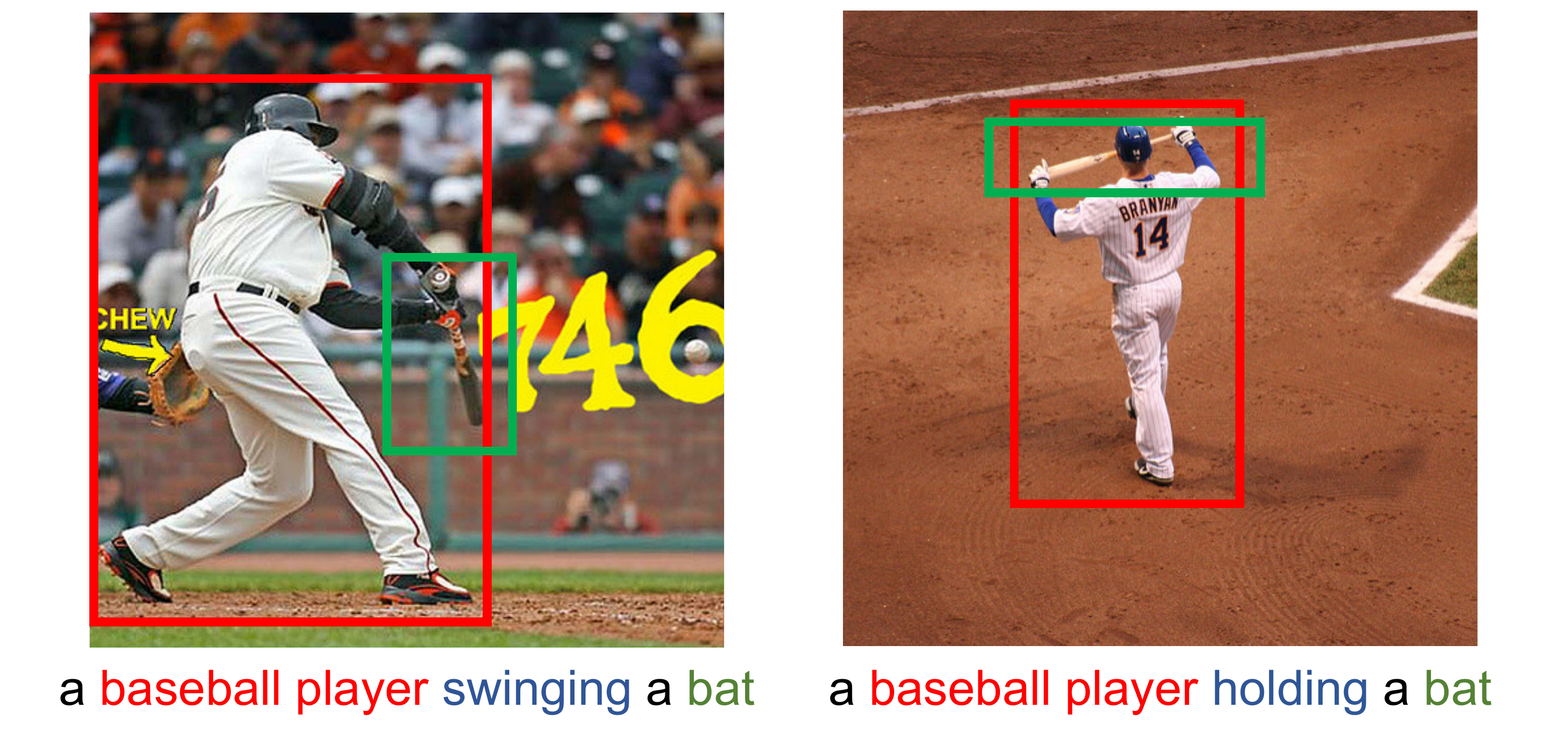}
\end{center}
\caption{Captions make references to objects and stuffs, and relations among them. In this figure, subjects, objects, relation predicates are respectively highlighted with red, green, blue. Image-text matching methods that only rely on objects (``player'' and ``bat'') and ignore relations (``holding'' and ``swinging'') would consider both captions are correct for both images. (Best view in color)}
\label{fig:concept}
\end{figure}

Despite these progress, vision-and-language remains challenging partly due to the fact that interplay between objects and stuffs is not taken into account.
Unlike attributes and actions (of single objects) that may be inferred from individual object/region features, visual relations are not considered at all in many state of the art models (e.g. the top-down captioner \cite{anderson2017bottom} and the SCAN model for image-text matching \cite{lee2018stacked}).
To understand the crux of the matter, we present two examples in Fig. \ref{fig:concept} where \textbf{``a baseball player swinging a bat''} and \textbf{``a baseball player holding a bat''} are captions of two different images.
However, say image-text matching models that do not consider visual relations (``holding'' and ``swinging'') could score both captions equally good for both images, thus fail to align the captions to their corresponding images.
The same issue can be found in models for other applications such as caption generation and visual question answering \cite{yao2018exploring}. 

Detecting visual relations between objects and stuffs is an emerging research problem that has drawn significant attention recently \cite{yang2018graph,zellers2018neural,dai2017detecting,li2017scene,lu2016visual,plummer2017phrase,xu2017scene,li2017vip,liang2017deep,newell2017pixels,peyre2017weakly,zhang2017visual,zhang2017ppr,zhuang2017towards} after the release of several large-scale visual-relation-labeled datasets such as visual genome \cite{krishna2017visual} and HICO \cite{chao2015hico}. 
In particular, researchers have developed scene graph generators by combining region and relationship detection models.
Given an image, scene graph generators predict relation triplets $<subject, predicate, object>$.
We hypothesize that training neural scene graph generators for relation detection, by necessity, would also learn embedding features with rich semantics of visual relations.
Assuming this is true, potentially these embedding features could provide prior knowledge of visual relations for various language-and-vision applications.
To empirically verify this hypothesis, we incorporate the scene graph generator features into state of the art models for image-text matching and image caption generation.
For image-text matching, we propose a new relation-based Stacked Cross Attention Network (R-SCAN) based on SCAN \cite{lee2018stacked}.
R-SCAN additionally encodes visual relations and employs a gating mechanism to adaptively select region and relation features.
Similarly, we extend the top-down captioner \cite{anderson2017bottom} with an additional attention component for relation features.

Previous scene graph generators are usually trained and evaluated on Visual Genome \cite{krishna2017visual} splits that consists of the most common visual relationships (e.g. VG150 dataset \cite{xu2017scene}).
However, such datasets are problematic in that they mainly contain common relations whose corresponding predicates can be easily detected using statistical counting based on the text context without the need of truly understanding visual relations \cite{zellers2018neural,liang2019rethinking}. 
For example, they would predict that the relation between ``a baseball player'' and ``a bat'' is most likely ``swing'' rather than ``throw'', because ``swing'' co-occurs more often with ``baseball player'' and ``bat'' in data.
In another word, scene graph generators learn to take the easy way out with such datasets.
In light of this, Liang \textit{et al.} \cite{liang2019rethinking}, in parallel to this work, created VrR-VG dataset which contain much richer categories of relations that cannot be easily detected based solely on statistical counting.
We therefore also resort to VrR-VG for training scene graph generators.

The experimental results show that R-SCAN significantly improves bi-directional retrieval metrics compared with SCAN which is the current state of the art (e.g. it improves recall@1 of image retrieval on Flickr30K \cite{young2014image} by 12.2\% relatively).
Our relation-based top-down captioner \cite{anderson2017bottom} also improves CIDEr score \cite{vedantam2015cider} from 113.5 to 114.9 and SPICE score \cite{anderson2016spice} from 20.3 to 20.9 on MSCOCO \cite{lin2014microsoft}.

A major difference between this work and recent works \cite{yao2018exploring,hou2019relational,yang2018scene} that also explore visual relations for vision-and-language is in that we do NOT use graph convolution networks (GCN). 
For example, Yao \textit{et al.} \cite{yao2018exploring} drew connections between regions with scene graph generators or proximity-based heuristics (whereas no semantic information attached to these connections) and built complicated graph convolution to implicitly capture visual relations from caption data.
We argue that directly transferring knowledge of visual relations from scene graph generators instead of discarding them is a much simpler and is an equivalently effective alternative to GCN.
We show that our approach is generic and applicable to metric learning (image-text matching) and sequence prediction (image caption generation) tasks.


\section{Related Work}
\label{sec:related_work}

\noindent\textbf{Image-Text Matching.}
The goal of image-text matching is learning similarity between images and text descriptions, and is usually evaluated on bi-directional image and text retrieval tasks.
There has been an extensive line of work addressing image-text matching using neural networks \cite{kiros2014unifying,vendrov2015order,ba2016layer,wang2016learning,klein2015associating,lev2016rnn,zheng2017dual,faghri2017vse++,peng2017cm,gu2017look,eisenschtat2017linking,devlin2015language,fang2015captions,karpathy2015deep,niu2017hierarchical,huang2017learning,huang2017instance,nam2016dual,lee2018stacked}.
In particular, R-SCAN model proposed in this paper is built on Stacked Cross Attention Network (SCAN) \cite{lee2018stacked} that uses a two-stage attention mechanism to discover fine-grained correspondence between objects/stuffs and words.
R-SCAN additionally encodes visual relations and employs a gating mechanism to select between region and relation features.

\noindent\textbf{Image Captioning.}
Image captioning refers to automatic image description generation and has also been widely studied over the years \cite{vinyals2015show,fang2015captions,xu2015show,rennie2017self,liu2017improved}.
Recently, Anderson \textit{et al.} \cite{anderson2017bottom} proposed ``bottom-up attention``, which refers to extracting and encoding salient regions of object and stuff bottom-up from perceptional priors that region detectors (e.g. Faster R-CNN) learn through pre-training.
Bottom-up attention dramatically improve various vision-and-language tasks including image captioning \cite{anderson2017bottom}.
We extend the top-down captioner proposed by Anderson \textit{et al.} \cite{anderson2017bottom}, adding relation features from scene graph generators along with region features from bottom-up attention to help the captioning model capture visual relations.

\noindent\textbf{Scene Graph Generation.}
The scene graph generation task has recently attracted significant interest from the vision community \cite{sadeghi2011recognition,divvala2014learning,yang2018graph,zellers2018neural,dai2017detecting,li2017scene,lu2016visual,plummer2017phrase,xu2017scene,li2017vip,liang2017deep,newell2017pixels,peyre2017weakly,zhang2017visual,zhang2017ppr,zhuang2017towards}. 
In most of these works, visual relationship is treated as edges between two objects in the scene graph, and many previously proposed approaches have used context propagation mechanism. 
Xu \textit{et al.} \cite{xu2017scene} presented an iterative message passing framework to predict object and their relationships jointly by using two separate networks, one for edge and one for nodes.
In \cite{zellers2018neural}, Zellers \textit{et al.} designed Stacked Motif Network to capture higher order substructures in scene graphs. 
Stacked Motif Network encodes each relation triplet $<subject, predicate, object>$ into an embedding vector which we use in this work as a source of visual relation prior for downstream applications.

\begin{figure*}[t!]
\begin{center}
\includegraphics[width=\textwidth]{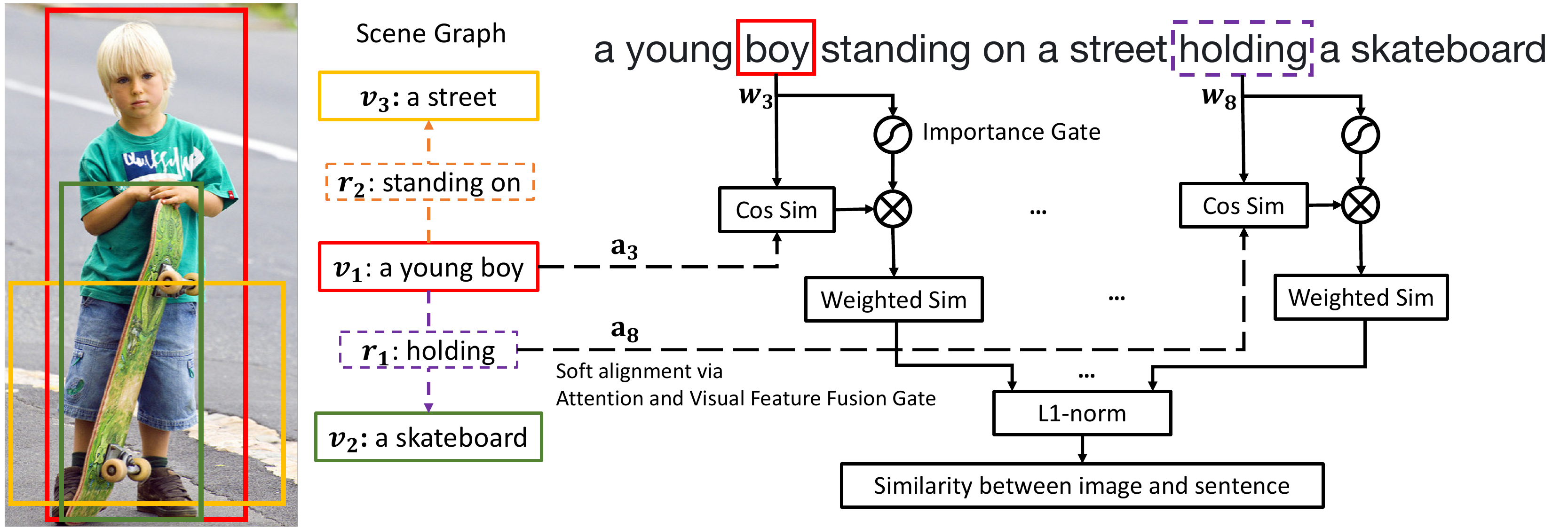}
\end{center}
\caption{Overview of the proposed R-SCAN model. A soft alignment mechanism using attention and visual feature fusion gate (selecting between region and relation features) aligns region features ($v_i$) or relation features ($r_l$) to words ($w_j$) (details explained in Sec. \ref{subsec:r_scan}). Cosine similarity (Cos Sim) is computed between each word and the corresponding aligned visual feature. We use an importance gate determined by each word to calibrate the importance of individual cosine similarity values in the final similarity between the whole image and the full sentence.}
\label{fig:r_scan}
\end{figure*}

Scene graph generators are usually trained and evaluated on Visual Genome \cite{krishna2017visual} splits dominated with the most common visual relationships (e.g. VG150 dataset \cite{xu2017scene}).
It is pointed out in \cite{zellers2018neural,liang2019rethinking} that such relation data would lead scene graph generators fitting to statistical counting based on textual context instead of truly understanding visual relations.
This finding implies existing scene graph generation benchmarks are potentially not ideal.

\noindent\textbf{Bringing Visual Relations to Vision-and-Language.} 
Johnson \textit{et al.} \cite{johnson2015image} proposed a framework using ground-truth scene graph as the query for image retrieval, but in practice it is still very difficult to construct accurate scene graphs from either text or images.
Several recent works proposed GCN-based models for image captioning and visual question answering \cite{hou2019relational,yang2018scene,yao2018exploring}.
For example, Yao \textit{et al.} \cite{yao2018exploring} designed a GCN-based captioning model that employs either heuristics of spatial proximity or scene graph generators to propose possible connections between objects (whereas no semantic information attached to these connections).
These approaches discard semantic labels and representations of visual relations coming from scene graph generators, and instead implicitly infer relationships from caption data.
Directly utilizing scene graph generator features avoids expensive graph convolution, and we argue it is still effective in capturing visual relations.
On the other hand, Liang \textit{et al.} \cite{liang2019rethinking} show that additional visual relation prediction objective enriches region features and improves downstream image captioning and visual question answering models, but their proposal does not actually model visual relations and thus is lack of explainability.


\section{Methods}
\label{sec:method}

Sec. \ref{subsec:r_scan} describes R-SCAN model which leverages visual relation features from scene graph generators to improve the image-text matching. 
In Sec. \ref{subsec:captioning}, we present the proposed relation-based top-down captioner which extends the top-down captioner from Anderson \textit{et al.} \cite{anderson2017bottom}.
Sec \ref{subsec:SGG} explains how we pre-train scene graph generators to learn effective relation features.

\subsection{R-SCAN for Image-Text Matching}
\label{subsec:r_scan}

The architecture of R-SCAN is presented in Fig. \ref{fig:r_scan}.
R-SCAN consists of three components: 
(1) a text encoder,
(2) a visual encoder for features of image region and visual relation, and
(3) an attention module for aligning image regions and visual relations to words and calculating image-text similarity.

\noindent\textbf{Text encoder.} The text encoder is identical to SCAN \cite{lee2018stacked}. It takes as input a sequence of $n$ words, each being represented as a one-hot vector, and maps each word into a 300-dimensional vector as
\begin{align}
\label{eq:wemb}
\mathbf{x_i} = \mathbf{W_e} \mathbf{\hat{w}_i} 
\end{align}
where $i \in \{1,2, \dots, n\}$, $\mathbf{W_e}$ is a randomly initialized embedding matrix and $\mathbf{\hat{w}_i}$ is the one-hot representation of the $i$-th word. 
We then use a bi-directional GRU to generate for each word $\mathbf{x_i}$ a contextual embedding vector by infusing contextual information from both sides of the word in the text.
The bi-directional GRU contains a forward GRU which reads the word sequence $T$ from left to right to produce the hidden states:
\begin{equation}
\overrightarrow{\mathbf{h_i}} = \overrightarrow{GRU}(\mathbf{x_i})
\end{equation}
and similarly a backward GRU which reads $T$ from right to left to produce the hidden states $\overleftarrow{\mathbf{h_i}}$.
The contextual embedding vector of word $\mathbf{w_i}$ is obtained by averaging the forward hidden state $\overrightarrow{\mathbf{h_i}}$ and backward hidden state $\overleftarrow{\mathbf{h_i}}$: 
\begin{equation}
\mathbf{w_i} = \dfrac{(\overrightarrow{\mathbf{h_i}} + \overleftarrow{\mathbf{h_i}})}{2}
\end{equation}


\noindent\textbf{Visual encoder.} We use a pre-trained Faster R-CNN (identical to SCAN \cite{lee2018stacked}) for extracting representations of object and stuff, denoted as $\{ \mathbf{\hat{v}_1}, \mathbf{\hat{v}_2}, \dots, \mathbf{\hat{v}_k} \}\ $ where $k$ is number of regions detected in an image.
On the other hand, we use a pre-trained Stacked Motif Networks (a scene graph generator proposed by Zellers \textit{et al.} \cite{zellers2018neural}) for extracting representations of visual relations $<subject, predicate, object>$, denoted as $ \{ \mathbf{\hat{r}_1}, \mathbf{\hat{r}_2}, \dots, \mathbf{\hat{r}_m} \}$ where $m$ is number of visual relations detected in an image.
$\mathbf{\hat{v}_i}$ and $\mathbf{\hat{r}_l}$ are subsequently transformed to $h$-dimensional vectors:
\begin{equation}
\mathbf{v_i} = W_v\mathbf{\hat{v}_i} + b_v
\end{equation}
\begin{equation}
\mathbf{r_l} = W_r\mathbf{\hat{r}_l} + b_r
\end{equation}

\noindent\textbf{Attention module for similarity inference.} The attention module generalizes the previously proposed SCAN t-i model \cite{lee2018stacked} and \textit{softly aligns}\footnote{Attention on visual representations (of region or relation) w.r.t. a word in text fuses visual representations deferentially based on their relevance to the word. Such process can be considered as a \textit{soft alignment} of relevant visual representations w.r.t. the word. We further introduce a ``visual feature fusion gate'' conditioned on the word to fuse the attended region and relation representation deferentially. This process is considered as a \textit{soft decision} of whether to align region or relation w.r.t. the word. } representations of region and relation in image with words in text and infer the similarity between image and text.

Given feature vector of regions $\mathbf{v}$, relations $\mathbf{r}$ and words $\mathbf{w}$, attention weights $att^{rel}$ and $att^{rgn}$ are computed as 

\begin{equation}
\label{eq:rel_att}
att^{rels}_{lj} = \frac{\exp(\lambda^{rel}\hat{s}_{lj}^{rel})}{\sum\limits_{l=1}^{m}\exp(\lambda^{rel}\hat{s}_{lj}^{rel})}
\end{equation}

\begin{equation}
\label{eq:reg_att}
\small
att^{rgn}_{ij} = \frac{\exp(\lambda^{rgn}\hat{s}_{ij}^{rgn})}{\sum\limits_{i=1}^{k}\exp(\lambda^{rgn}\hat{s}_{ij}^{rgn})}
\end{equation}

\noindent
where $\lambda^{rgn}$ and $\lambda^{rel}$ are temperature hyper-parameters \cite{chorowski2015attention}. 
Following \cite{lee2018stacked}, the similarity between $l$-th relation and $j$-th word $\hat{s}_{lj}^{rel}$ is computed as

\begin{equation}
s_{lj}^{rel}=\frac{\mathbf{r_l}^T\mathbf{w_j}}{\lVert \mathbf{r_l} \rVert \lVert \mathbf{w_j} \rVert}
\end{equation}

\begin{equation}
\hat{s}_{lj}^{rel} = \frac{[s_{lj}^{rel}]_+}{\sqrt[]{\sum_{l=1}^m [s_{lj}^{rel}]_+^2}}
\end{equation}

\noindent
where $[x]_+ \equiv max(x,0)$. The similarity between $i$-th region and $j$-th word $\hat{s}_{ij}^{rgn}$ is computed as

\begin{equation}
s_{ij}^{rgn}=\frac{\mathbf{v_i}^T\mathbf{w_j}}{\lVert \mathbf{v_i} \rVert \lVert \mathbf{w_j} \rVert}
\end{equation}

\begin{equation}
\hat{s}_{ij}^{rgn} = \frac{[s_{ij}^{rgn}]_+}{\sqrt[]{\sum_{i=1}^k [s_{ij}^{rgn}]_+^2}}
\end{equation}


Given attention weights $att^{rel}_{ij}$, the \emph{attended} relation representation of the image w.r.t. word $\mathbf{w_j}$ is defined as

\begin{equation}
\mathbf{a^{rel}_j} = \sum\limits_{l=1}^{m} att^{rel}_{lj}\mathbf{r_l}
\end{equation}

\noindent
where $\mathbf{a^{rel}_j}$ can be viewed as a summarized relation vector of the image generated using a fusing process where all the relation vectors are weighted by their attention weights of equation \eqref{eq:rel_att} w.r.t. $\mathbf{w_j}$ and aggregated.
$\mathbf{a^{rel}_j}$ can also be considered representing the \textit{soft alignment} between $\mathbf{w_j}$ and the relations in the image. A special case of the soft alignment is \emph{hard} alignment where there is only one relation that has a non-zero attention weight w.r.t. $\mathbf{w_j}$.

Similarly, given attention weights $att^{rgn}_{ij}$, the attended region representation of the image w.r.t. word $\mathbf{w_j}$ is defined as:

\begin{equation}
\label{eq:attended_objs_rels}
\mathbf{a^{rgn}_j} = \sum\limits_{i=1}^{k} att^{rgn}_{ij}\mathbf{v_i}\\
\end{equation}


We now define the attended representation of the image w.r.t. word $\mathbf{w_j}$, denoted as $\mathbf{a_j}$, which combines $\mathbf{a^{rel}_j}$ and $\mathbf{a^{rgn}_j}$.
Considering that while entities/nouns often attend to objects and stuffs in an image, predicates to relations, we introduce a \emph{visual feature fusion gate} that conditions on each word (type) to fuse $\mathbf{a^{rel}_j}$ and $\mathbf{a^{rgn}_j}$ using a mixture model:

\begin{equation}
\label{eq:visfeagate}
g_{vf}(\mathbf{w_j}) = \sigma(\mathbf{\omega}^{T}_{vf}\mathbf{w_j} + \beta_{vf})
\end{equation}
\begin{equation}
\mathbf{a_j} = g_{vf}(\mathbf{w_j})\mathbf{a^{rel}_j} + (1 - g_{vf}(\mathbf{w_j}))\mathbf{a^{rgn}_j}
\end{equation}

\noindent
where $ \sigma(.) $ is the sigmoid function, $\beta_{vf}$ is bias and $\mathbf{\omega}_{vf}$ is a trainable projection vector with the same dimension as $\mathbf{w_j}$. 

\noindent
The similarity between the whole image and $j$-th word is computed as: 
\begin{equation}
\label{eq:cosine}
R(\mathbf{a_j}, \mathbf{w_j}) = \frac{\mathbf{a_j}^T\mathbf{w_j}}{\lVert \mathbf{a_j} \rVert \lVert \mathbf{w_j} \rVert}
\end{equation}

Finally, we need to compute the similarity between the image and the text which consists of a set of words. In SCAN t-i \cite{lee2018stacked}, this is achieved by averaging or LogSumExp pooling the word-image similarity over all the words in the text. 
In R-SCAN, we assign each word an importance weight using a machine-learned importance gate, similar to the visual feature fusion gate of equation \eqref{eq:visfeagate}:

\begin{equation}
g_{impt}(\mathbf{w_j}) = \sigma(\omega^{T}_{impt}\mathbf{w_j} + \beta_{impt})
\end{equation}

The similarity between image $V$ and text $T$ is defined as the sum of the $\ell_1$ norm of the weighted word-image similarity over all the words in the text:
\begin{equation}
\label{eq:global_sim}
sim(V,T) = \sum\limits_{j=1}^{n} {\lVert \hat{R}(\mathbf{a_j}, \mathbf{w_j}) \rVert}_1
\end{equation}
\begin{equation}
\hat{R}(\mathbf{a_j}, \mathbf{w_j}) = g_{impt}(\mathbf{w_j})R(\mathbf{a_j}, \mathbf{w_j})
\end{equation}


\noindent

\noindent
\textbf{Learning objective}. Following \cite{lee2018stacked,faghri2017vse++}, we use hinged-based triplet ranking loss and focus on the hardest negatives in a mini-batch. 
For a positive pair $(V, T)$, we generate two negative pairs by picking a mismatched image 
$V^- = argmax_{V'\neq V}sim(V',T)$ 
and a mismatched text
$T^- = argmax_{T'\neq T}sim(V,T')$, respectively. 
The loss function is defined as
\begin{equation}
\begin{aligned}
l(V, T) &= [\alpha - sim(V,T) + sim(V,T^-]_{+} \\
        &+ [\alpha - sim(V,T) + sim(V^-,T)]_+
\end{aligned}
\label{triplet_loss_hard}
\end{equation}
where $[x]_+ \equiv max(x,0)$ and $\alpha$ is the margin, which in this work is set to $0.2$.

\begin{figure}[t!]
\begin{center}

\includegraphics[width=1.0\linewidth]{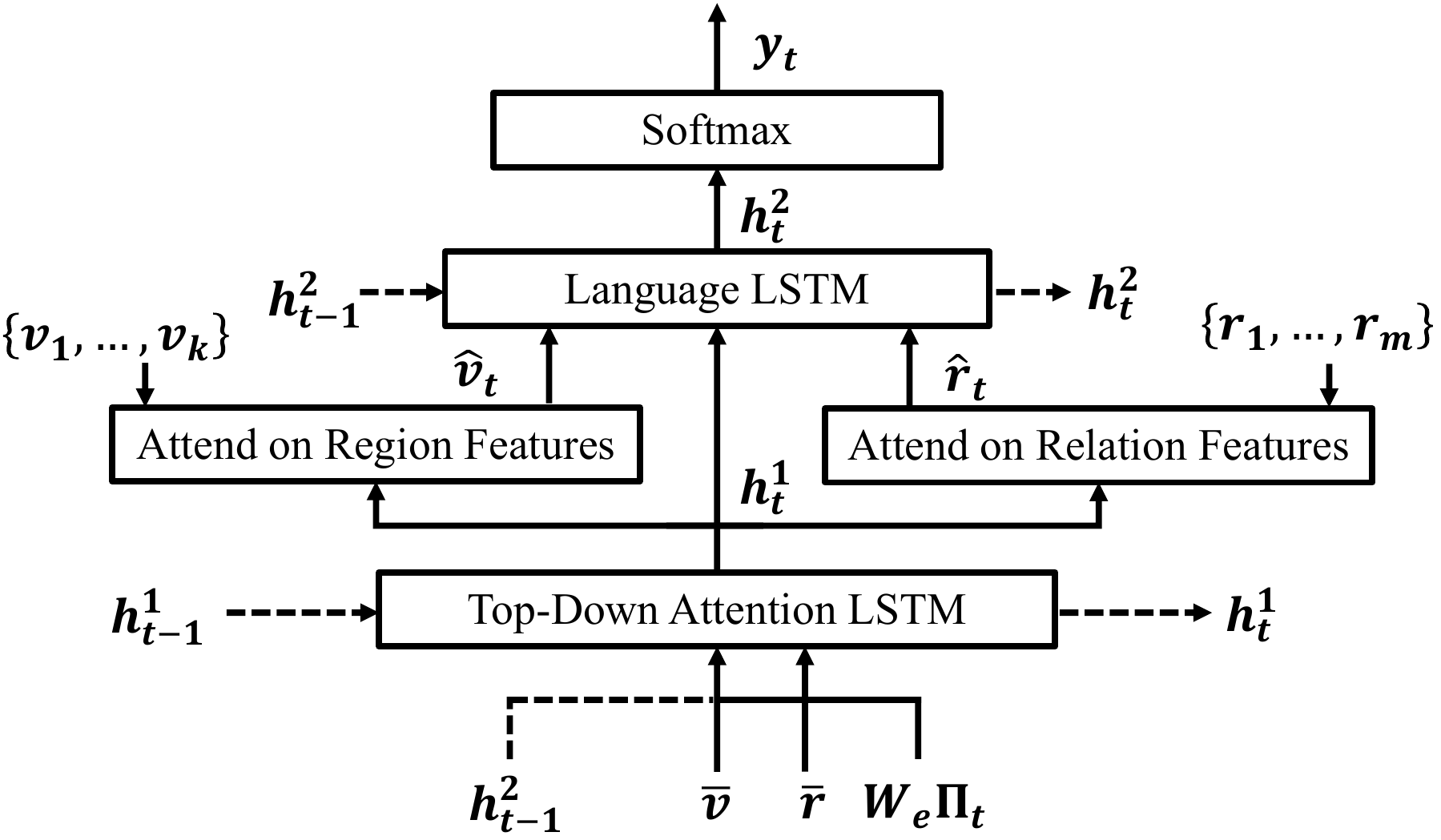}
\end{center}
\caption{Overview of the proposed captioning model. Two LSTM layers are used to selectively attend to image region features extracted from Faster R-CNN ${\mathbf{v_1}, \dots, \mathbf{v_k}}$ and relation features extracted from Stacked Motif Networks ${\mathbf{r_1}, \dots, \mathbf{r_m}}$.}
\label{fig:caption_model}
\end{figure}

\subsection{Relation-based Top-Down Captioner}
\label{subsec:captioning}

For image captioning, we propose a simple extension of the top-down captioner \cite{anderson2017bottom}, adding relation features from Stacked Motif Networks as shown in Fig. \ref{fig:caption_model}.
Most parts of the model definition are identical to the original as described in Sec 3.2 and Fig. 3 of \cite{anderson2017bottom}.
To include relation features, we change the input vector to the attention LSTM at each time step to concatenation of the mean-pooled relation feature $\mathbf{\bar{r}} = \frac{1}{m}\sum\nolimits_{l=1}^m \mathbf{r}_l$, the mean-pooled image region feature $\mathbf{\bar{v}} = \frac{1}{k}\sum\nolimits_{i=1}^k \mathbf{v}_i$, the previous output of the language LSTM ($\mathbf{h}^2_{t-1}$), and an encoding of the previously generated word ($W_e\Pi_t$) \cite{anderson2017bottom}. 
The attended relation feature $\mathbf{\hat{r}}_t$ is obtained in the same way as attended image feature $\mathbf{\hat{v}}_t$. $\mathbf{\hat{r}}_t$ is then concatenated to the input to the language model LSTM, in addition to the attended image feature $\mathbf{\hat{v}}_t$ and the output of the attention LSTM ($\mathbf{h}^1_t$).
The learning objectives for cross entropy training and subsequent self-critical CIDEr optimization \cite{rennie2017self} are identical to the top-down captioner \cite{anderson2017bottom}.
We also employ the same Faster R-CNN model for bottom-up attention.

\subsection{Scene Graph Generator as Feature Learner}
\label{subsec:SGG}

In our framework, relation features are extracted from neural scene graph generators.
Specifically, we uses Stacked Motif Network \cite{zellers2018neural} as the default scene graph generator for all the experiments.
Stacked Motif Networks predict graph elements by staging bounding box predictions, object classifications, and relationships such that the global context encoding of all previous stages establishes rich context for predicting subsequent stages.
We take the 4096-d relation representation before applying the final projection and softmax function to represent relation triplets $<subject, predicate, object>$ (see Sec. 4.3 of \cite{zellers2018neural}).

Previous scene graph generators are usually trained and evaluated on Visual Genome \cite{krishna2017visual} splits that consists of the most frequent visual relationships in Visual Genome.
VG150 dataset is one of the most used benchmark \cite{xu2017scene}, but this data could be problematic because it consists of most frequent 50 relation predicates and 150 object categories in Visual Genome, and these common relation predicates in VG150 can often be detected statistical counting without understanding of the images \cite{zellers2018neural,liang2019rethinking}.
As a result, although the scene graph generators developed on VG150 are often reported to perform well on the VG150 test set, the high performance does not translate to visible gains in end applications such as captioning and visual question answering, as reported in Sec. \ref{sec:experiments} and in \cite{liang2019rethinking}.
Other similar benchmarks also suffer from the same cause.

In light of this, we resort to training Stacked Motif Networks with \textbf{VrR-VG} dataset \cite{liang2019rethinking}. 
VrR-VG was created by choosing a subset of Visual Genome and removing the predicates that can be easily predicted solely using language models. 
We carefully avoid training data contamination, excluding any images that are in MSCOCO validation and test sets (created by Karpathy \textit{et al.} \cite{karpathy2015deep}) from our training split. 

In this work, we do not draw relations between visual relation detection results and numbers on end applications, although we did find that features from \cite{zellers2018neural} lead to better results than \cite{xu2017scene} whose results on VG150 is inferior.
First of all, as mentioned in Sec. \ref{sec:related_work}, it has been found that VG150 and similar benchmarks might not be ideal for visual relations \cite{zellers2018neural,liang2019rethinking}.
Secondly, the common mAP metric becomes problematic with datasets like VrR-VG that contain large number of object and/or relation classes.
As most of the classes are in tail of distributions, they can barely be accurately predicted yet mAP takes average of per-class precision.

We would also like to clarify that using additional relation features do not mean including additional training images.
We pre-train Stacked Motif Networks with VrR-VG or VG150 which are subsets of Visual Genome, whereas the baseline methods SCAN \cite{lee2018stacked} and top-down captioner \cite{anderson2017bottom} also use Visual Genome for pre-training of bottom-up attention Faster R-CNN models.

\section{Experiments}
\label{sec:experiments}

\noindent\textbf{Datasets.} We evaluate R-SCAN on the MSCOCO \cite{lin2014microsoft} and Flickr30K \cite{young2014image} datasets.
Relation-based top-down captioner is only evaluated on MSCOCO following prior work.
Flickr30K contains 31,000 images collected from Flickr with five captions each. Following the MSCOCO splits that Andrej Karpathy created \cite{karpathy2015deep,faghri2017vse++}, we use 1,000 images for validation and 1,000 images for testing and the rest for training. 
MSCOCO contains 123,287 images, and each image is annotated with five text descriptions. 
In \cite{karpathy2015deep}, the dataset is split into 82,783 training images, 5,000 validation images and 5,000 test images.
We follow \cite{faghri2017vse++,lee2018stacked} to add 30,504 images that were originally in the validation set of MSCOCO but have been left out in this split into the training set. 
Each image comes with 5 captions. 
The results are reported on full 5K test images or averaging over 5 folds of 1K test images. 
As is common in information retrieval, we measure performance of sentence retrieval (image query) and image retrieval (sentence query) by recall at $K$ (r@$K$) defined as the fraction of queries for which the correct item is retrieved in the closest $K$ points to the query.
Also following prior work, we evaluate captioning with CIDEr score \cite{vedantam2015cider} which captures the syntactic correctness and SPICE score \cite{anderson2016spice} which reflects whether our models generate right descriptions of scene.

\noindent\textbf{Implementation details.} As mentioned in Sec. \ref{subsec:SGG}, we use Stacked Motif Networks to learn relation features. 
Top $m$ relation features are chosen based on the triplet confidence score following \cite{zellers2018neural}.
We fix $m$ to 36 for the following experiments but using $m=18$ can result in similar performance in most of the cases.
Increasing $m$ to 72 results in performance degradation due to noisy information. Stacked Motif Networks pre-trained with VG150 that we used in experiments is available publicly\footnote{https://github.com/rowanz/neural-motifs}. 
We train Stacked Motif Networks on VrR-VG and matches the results reported in \cite{liang2019rethinking} for object detector, scene graph classification and scene graph detection. 
The object detector for Stacked Motif Networks is selected to be Faster R-CNN with VGG backbone \cite{simonyan2014very}. 
We have also experimented with ResNet-101 backbone \cite{he2016deep} but did not observe difference in performance.

To detect and encode image regions, we adopt the same Faster R-CNN model from \cite{anderson2017bottom} as our bottom-up attention model.
Top 36 regions were selected per image following the same criterion in \cite{anderson2017bottom,lee2018stacked}. 
Although region features from Stacked Motif Networks can also be used, we choose the bottom-up attention model for fair comparison with SCAN \cite{lee2018stacked} and top-down captioner \cite{anderson2017bottom}.

For R-SCAN, softmax temperature $\lambda^{rel}$ and $\lambda^{rgn}$ are selected on the validation set. We use Adam optimizer \cite{kingma2014adam} to train the models. 
R-SCAN models are trained with a learning rate of 0.0005 for 10 epochs and then 0.00005 for another 10 epochs, following SCAN \cite{lee2018stacked}.
For captioning, we follow the training and evaluation configurations \cite{anderson2017bottom}.

\begin{figure*}[t!]
\begin{center}
\includegraphics[width=1.0\linewidth]{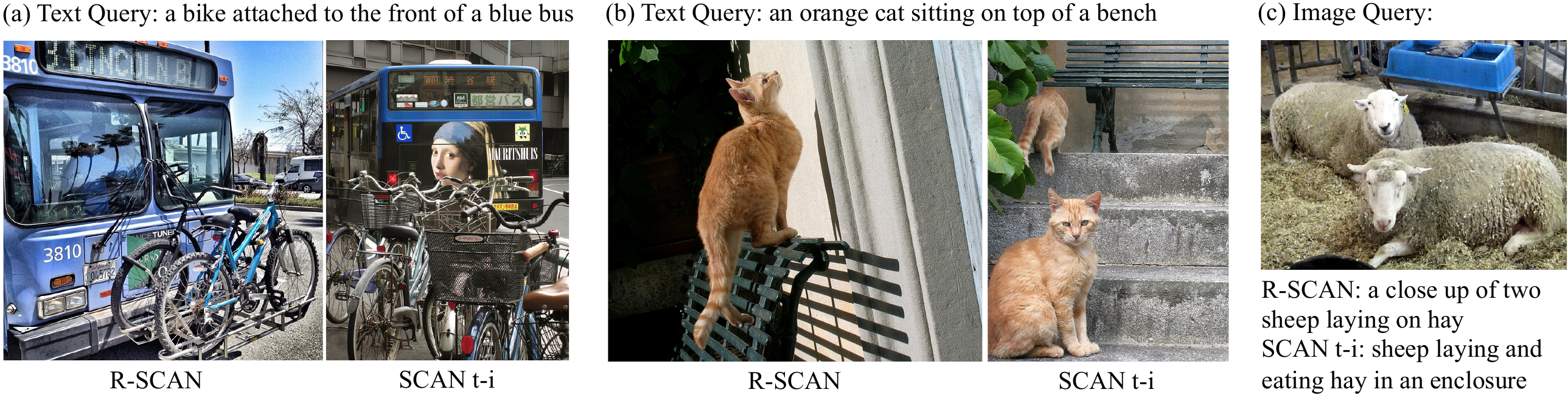}
\end{center}
   \caption{(a)(b) are qualitative examples of image retrieval given text queries using R-SCAN and SCAN t-i AVG on (COCO-test-VrR. We show the top-1 ranked images. In (a) it can be observed that the predicate ``attached to the front of'' makes the difference between R-SCAN and SCAN's results as objects ``bike'' and ``bus'' present in both images. Similarly, in (b) both ``cat'' and ``bench'' present in image, but SCAN does not capture the relation ``sitting on top of''. (c) is an example of text retrieval given image query, where SCAN incorrectly ignores ``eating'' and R-SCAN captures the relation ``laying on''.}
\label{fig:rel_qual}
\end{figure*}

\subsection{The Effectiveness of Visual Relations}

In the following analysis, we investigate the quality of image-text matching specifically on captions that describe visual relations and corresponding images, and compare Stacked Motif Network features pre-trained on VrR-VG and VG150.
The motivation is to focus only on relation-relevant predicts to best quantify the improvements coming from relation features.
Zellers \textit{et al.} \cite{zellers2018neural} analyzed Visual Genome dataset and concluded that the predominant relations are geometric (above, behind, under) and possessive (has, part of, wearing). Such relations are often obvious, e.g., houses tend to have windows. 
VrR-VG dataset rules out relations that could be easily predicted with language prior, and clusters the remaining high frequency relations based on semantic similarity to 117 predicates \cite{liang2019rethinking}. 
By comparing VrR-VG and the original Visual Genome, those 117 relations can be mapped back to 259 relation predicates in Visual Genome, where 164 of them are identified by us as semantic relations (leaning on, walking towards, jumping on) which correspond to activities, are less frequent and less obvious (definition of semantic visual relations can be found in \cite{zellers2018neural}). 
We found that there are 3,403 images in MSCOCO 5K test set with at least one ground truth caption that has one of the 164 semantic predicates.
We use those images and randomly sample one corresponding caption that describes visual relations to construct a new \textbf{COCO caption test split with visually relevant relations (COCO-test-VrR)} which allows us to focus on improvements of image-text matching that involves visual relations.

\begin{table}[t!]
\footnotesize
\begin{center}
\begin{tabular}{l|cccccc}
\hline
\noalign{\smallskip}
& \multicolumn{3}{c}{text-to-image} & \multicolumn{3}{c}{image-to-text} \\
Model & r@1 & r@5 & r@10 & r@1 & r@5 & r@10 \\
\hline
SCAN t-i AVG & 37.9 & 69.4 & 80.8 & 38.5 & 70.7 & 82.5 \\
R-SCAN-VG150 & 39.8 & 70.6 & 82.0 & 38.1 & 71.0 & 83.5 \\
R-SCAN-VrRVG & 40.1 & 70.5 & 81.8 & 39.6 & 72.7 & 83.7 \\
\hline
\end{tabular}
\end{center}
\caption{Comparison of the cross-model retrieval results in terms of recall@K (r@K) on COCO-test-VrR. `text-to-image' denotes image retrieval given text query. `image-to-text' denotes text retrieval given image query.}
\label{tb:r_coco}
\vspace{-0.5em}
\end{table}

\begin{table*}[t!]
\small
\begin{center}

\begin{tabular}
{p{4.2cm}p{0.6cm}p{0.6cm}p{0.7cm}p{0.6cm}p{0.6cm}p{0.7cm}p{0.6cm}p{0.6cm}p{0.7cm}p{0.6cm}p{0.6cm}p{0.7cm}}
\hline
\noalign{\smallskip}
& \multicolumn{6}{c}{Flickr30K 1K Test Images} & \multicolumn{6}{c}{MSCOCO 5-fold 1K Test Images} \\
\noalign{\smallskip}
& \multicolumn{3}{c}{text-to-image} & \multicolumn{3}{c}{image-to-text} & \multicolumn{3}{c}{text-to-image} & \multicolumn{3}{c}{image-to-text} \\
Method & r@1 & r@5 & r@10 & r@1 & r@5 & r@10 & r@1 & r@5 & r@10 & r@1 & r@5 & r@10 \\
\noalign{\smallskip}
\hline
UVS \cite{kiros2014unifying} & 16.8 & 42.0 & 56.5 & 23.0 & 50.7 & 62.9 & - & - & - & - & - & - \\ 
DVSA \cite{karpathy2015deep} & 15.2 & 37.7 & 50.5 & 22.2 & 48.2 & 61.4 & 27.4 & 60.2 & 74.8 & 38.4 & 69.9 & 80.5 \\ 
HM-LSTM \cite{niu2017hierarchical} & 27.7 & - & 68.8 & 38.1 & - & 76.5 & 36.1 & - & 86.7  & 43.9 & - & 87.8 \\ 
DAN \cite{nam2016dual} & 39.4 & 69.2 & 79.1 & 55.0 & 81.8 & 89.0 & - & - & - & - & - & - \\ 
VSE++ \cite{faghri2017vse++} & 39.6 & 70.1 & 79.5 & 52.9 & 80.5 & 87.2 & 52.0 & 84.3 & 92.0 & 64.6 & 90.0 & 95.7 \\ 
Picturebook \cite{kiros2018illustrative} & - & - & - & - & - & - & 55.2 & 87.2 & 94.4 & 63.4 & 90.3 & 96.5 \\ 
GXN \cite{gu2017look} & 41.5 & - & 80.1 & 56.8 & - & 89.6 & 56.6 & - & 94.5 & 68.5 & - & 97.9 \\ 
SCO \cite{huang2017learning} & 41.1 & 70.5 & 80.1 & 55.5 & 82.0 & 89.3 & 56.7 & 87.5 & \textbf{94.8} & 69.9 & 92.9 & 97.5 \\ 
\hline
SCAN: \\
\textit{SCAN ensemble}$^\dagger$ \cite{lee2018stacked} & \textit{48.6} & \textit{77.7} & \textit{85.2} & \textit{67.4} & \textit{90.3} & \textit{95.8} & \textit{58.8} & \textit{88.4} & \textit{94.8} & \textit{72.7} & \textit{94.8} & \textit{98.4} \\ 
SCAN i-t AVG \cite{lee2018stacked} & 44.0 & 74.2 & 82.6 & \textbf{67.7} & 88.9 & 94.0 & 54.4 & 86.0 & 93.6 & 69.2 & 93.2 & 97.5 \\ 
SCAN t-i AVG \cite{lee2018stacked} & 45.8 & 74.4 & 83.0 & 61.8 & 87.5 & 93.7 & 56.4 & 87.0 & 93.9 & \textbf{70.9} & \textbf{94.5} & 97.8  \\ 
\hline
Ours:  \\ 
R-SCAN (VrR-VG) & \textbf{51.4} & \textbf{77.8} & \textbf{84.9} & 66.3 & \textbf{90.6} & \textbf{96.0} & \textbf{57.6} & \textbf{87.3} & 93.7 & 70.3 & \textbf{94.5} & \textbf{98.1} \\ 
\hline
\end{tabular}
\end{center}
\caption{The cross-modal retrieval results of R-SCAN in terms of recall@$K$(r@$K$) on Flickr30K 1K test set and MSCOCO 5-fold 1K test set comparing to the baseline SCAN t-i model and other prior works. `text-to-image' denotes image retrieval given text query. `image-to-text' denotes text retrieval given image query. Best numbers with single model are bolded. $^\dagger$: The SCAN ensemble is listed here only as a reference.}
\label{tb:comp_prior_work}
\end{table*}

\begin{table}[t!]
\footnotesize
\begin{center}
\begin{tabular}
{lcccccc}
\hline
\noalign{\smallskip}
& \multicolumn{6}{c}{MSCOCO 5K Test Images} \\
\noalign{\smallskip}
& \multicolumn{3}{c}{text-to-image} & \multicolumn{3}{c}{image-to-text} \\
Method & r@1 & r@5 & r@10 & r@1 & r@5 & r@10 \\
\noalign{\smallskip}
\hline
DVSA \cite{karpathy2015deep} & 10.7 & 29.6 & 42.2 & 16.5 & 39.2 & 52.0 \\ 
VSE++ \cite{faghri2017vse++} & 30.3 & 59.4 & 72.4 & 41.3 & 71.1 & 81.2 \\ 
GXN \cite{gu2017look} & 31.7 & - & 74.6 & 42.0 & - & 84.7 \\ 
SCO \cite{huang2017learning} & 33.1 & 62.9 & 75.5 & 42.8 & 72.3 & 83.0 \\
\hline
SCAN: \\
\textit{SCAN ens}$^\dagger$ \cite{lee2018stacked} & \textit{38.6} & \textit{69.3} & \textit{80.4} & \textit{50.4} & \textit{82.2} & \textit{90.0} \\
SCAN t-i AVG \cite{lee2018stacked} & 34.4 & 63.7 & 75.7 & \textbf{46.4} & 77.4 & 87.2 \\ 
\hline
Ours:  \\ 
R-SCAN (VrR-VG) & \textbf{36.2} & \textbf{65.5} & \textbf{76.7} & 45.4 & \textbf{77.9} & \textbf{87.9} \\ 
\hline

\end{tabular}
\end{center}
\caption{Comparison of the cross-modal retrieval results in terms of recall@$K$(r@$K$) on MSCOCO 5K test set. `text-to-image' denotes image retrieval given text query. `image-to-text' denotes text retrieval given image query. Best numbers with single model are bolded. $^\dagger$: \textit{SCAN ens} denotes SCAN ensemble, which is listed here only as a reference.}
\label{tb:coco_5k}
\end{table}

In Table \ref{tb:r_coco}, we report the results of the baseline SCAN t-i AVG model and R-SCAN trained on MSCOCO and evaluated on COCO-test-VrR. 
We consider R-SCAN models with relation features pre-trained on VG150 and VrR-VG. 
Comparing to the SCAN t-i baseline, it can be observed that improvements on bi-directional retrieval with R-SCAN-VG150 is limited.
Pre-training with VrR-VG (R-SCAN-VrRVG), on the other hand, leads to significant improvements.
The hypothesis is that VG150 majorly contains relations whose corresponding predicts can be easily predicted with statistical counting and thus does not require genuine visual understanding, while VrR-VG preserves semantically valuable relations that cannot be inferred solely from counting and therefore learning on VrR-VG requires forming features that are truly embedded with visually relevant information.
Based on this finding, we choose VrR-VG to train relation features for all the following experiments in this work.
In Fig. \ref{fig:rel_qual}, we present qualitative examples of image-text bi-directional retrievals using R-SCAN and the baseline SCAN t-i AVG model.


\subsection{Cross-Modal Retrieval Results}

In Table \ref{tb:comp_prior_work}, we compare R-SCAN with the baseline SCAN t-i AVG model as well as other state of the art methods on Flickr30K and MSCOCO (tested on 5-fold 1K test set). 
On Flickr30K, R-SCAN achieves the best single model image retrieval with recall@1 at 51.4. 
Comparing to SCAN i-t AVG, the relative improvement is 12.2\%. 
The R-SCAN model even outperforms SCAN ensemble on Flickr30K.
On MSCOCO 5-fold test set, R-SCAN achieves the best recall@1 at 57.6 for image retrieval (single model). 
Table \ref{tb:coco_5k} presents the results on the full MSCOCO 5K test set.
R-SCAN achieves the better performance than all previous single model on most metrics of cross-modal retrieval. 

It can be observed that the relative improvement on image retrieval is more significant than on text retrieval. 
We hypothesize that the underlying causes are the composition of Flickr30K and MSCOCO test sets and the recall@$K$ metric definition: as opposed to image retrieval where only one ground truth image exists for each text query, in text retrieval each image query corresponds to five ground truth captions. 
Any of them could count as a correctly retrieved item. However, not all of the five captions describe visual relations in the image. For example, ``a male in a blue shirt and a laptop and couch'' and ``a man is sitting on a couch with a dog using a laptop'' are both ground truth captions for an image in MSCOCO, but the former caption can be retrieved without understanding of semantic visual relations between the man and other major objects (e.g. sitting in a couch). 
In MSCOCO 5K test set, 68.0\% of the images correspond to at least one caption that has one of the 164 COCO-test-VrR predicates. 
Nonetheless, only 2.8\% of the images have five of such captions, despite that predicates could be rephrased in other captions and may not fall in the range of COCO-test-VrR predicates. 
Another observation that supports our hypothesis is that R-SCAN actually shows similar improvements on image and text retrieval on COCO-test-VrR (as shown in Table \ref{tb:r_coco}) where there is only one ground truth caption per image.

\subsection{Image Captioning on MSCOCO}
\label{subsec:coco_caption_result}

\begin{table}[t]
\small
\begin{center}
\begin{tabular}{lcc|cc}
& \multicolumn{2}{c}{Cross Entropy} & \multicolumn{2}{c}{CIDEr optim.} \\
\hline
Model & CIDEr & SPICE & CIDEr & SPICE \\
\hline
Top-down \cite{anderson2017bottom} & 113.5 & 20.3 & 120.1$^\dagger$ & 21.4$^\dagger$ \\
Top-down (reimpl.) & 113.8 & 20.6 & 125.5 & 21.6 \\
Ours & 114.9 & 20.9 & 126.1 & 21.8 \\
\hline
GCN-LSTM$_{spa}$ \cite{yao2018exploring} & 115.6 & 20.9 & 127.0$^\dagger$ & 21.9$^\dagger$ \\
GCN-LSTM$_{sem}$ \cite{yao2018exploring} & 116.3 & 20.9 & 127.6$^\dagger$ & 22.0$^\dagger$ \\
\hline
\end{tabular}
\end{center}
\caption{Image captioning performance in terms of CIDEr \cite{vedantam2015cider} and SPICE \cite{anderson2016spice} on the MSCOCO test split from \cite{karpathy2015deep}. Top-down (reimpl.) denotes our implementation of the top-down captioner. CIDEr optimization is performed 30 epochs for the re-implemented top-down captioner and our captioning models. $^\dagger$: Results are not directly comparable (refer to Sec. \ref{subsec:coco_caption_result} for details).}
\label{tb:coco_caption}
\end{table}

Table \ref{tb:coco_caption} shows the image captioning results of the proposed relation-based top-down captioner and baseline top-down captioner \cite{anderson2017bottom} on MSCOCO.
We report our results of optimizing the model for cross-entropy loss and subsequent policy gradient fine-tuning using CIDEr scores as rewards, following \cite{anderson2017bottom}.
Compared with the top-down captioner, our cross entropy model improves CIDEr score \cite{vedantam2015cider} from 113.5 to 114.9 and SPICE score \cite{anderson2016spice} from 20.3 to 20.9. 
We also present the results reported by Yao \textit{et al.} (GCN-LSTM) \cite{yao2018exploring} which exploits complicated GCN.

Interestingly, we found fine-tuning the original top-down captioner \cite{anderson2017bottom} with self-critical CIDEr optimization for 120 epochs (several days on single GPU), rather than training for less than one hour as reported in the original paper, can significantly boost CIDEr from 120.1 to 126.9 and SPICE from 21.4 to 21.8 (as on-policy reinforcement learning algorithms can take many epochs to converge).
This finding suggests that the documented results of the baseline in \cite{anderson2017bottom} are not comparable with the models whose training takes much more epochs. 
For example, the performance gap between \cite{anderson2017bottom} and \cite{yao2018exploring} might not be as large as indicated by the results reported in the two original papers, respectively.
For the sake of fairness, we compare the bottom-up baseline with our models, with both being optimized for CIDEr after 30 epochs. 
It can be observed in Table \ref{tb:coco_caption} that using relation features improves CIDEr score from 125.5 to 126.1 and SPICE score from 21.6 to 21.8.
The corresponding qualitative examples are presented in appendix (Fig. \ref{fig:caption_vis}).
The results show that, without GCN, our method is still effective in capturing visual relationships and improving image captioning.

\section{Conclusion}
\label{sec:conclusion}

In this study, we explored learning visual relation features for image-text matching and image caption generation with neural scene graph generators. 
By additionally capturing interplay between objects and stuffs, the proposed R-SCAN model achieves new state of the art result on the task of image-text cross-modal retrieval on the Flickr30K and MSCOCO benchmarks.
Similarly, relation-based top-down captioner also significantly improves image captioning.
The scene graph generator features are indeed effective in helping downstream models ground language to visual relations, but the crux of matters lies in pre-training scene graph generators with visually relevant relation data.
We hope this work would shed lights on the connection between scene graph generators and vision-and-language, and facilitate future research.

\section*{Acknowledgement} The authors thank Arun Sacheti and Pengchuan Zhang for their thoughtful feedback and discussions.

\clearpage

\appendix
\section*{Appendix Overview}
The supplementary material is structured as follows. 
Sec. \ref{r_coco_details} presents in details how the COCO-test-VrR test set is constructed. 
Sec. \ref{more_r_coco_qual} presents additional qualitative examples of cross-modal retrieval between image and text to demonstrate the effectiveness of R-SCAN and the use of visual relations for image-text matching. 
We also present image captioning examples to qualitatively demonstrate the effectiveness of using visual relations for the task.

\section{COCO-test-VrR}
\label{r_coco_details}

COCO-test-VrR is a subset of MSCOCO Karpathy 5K test split \cite{karpathy2015deep} introduced in Sec. 4.1 of the main paper. 
COCO-test-VrR focuses the evaluation of image-text matching on the captions that describe semantic visual relations \cite{zellers2018neural} and the corresponding images. 
We describe in detail how COCO-test-VrR is constructed in this section.

Zellers \textit{et al.} \cite{zellers2018neural} and Liang \textit{et al.} \cite{liang2019rethinking} have shown that a majority of the prevalent visual relations in Visual Genome \cite{krishna2017visual} could be predicted without visual information. 
Liang \textit{et al.} \cite{liang2019rethinking} constructed the Visually-Relevant Relationship Dataset (VrR-VG) which excludes the relations that could be easily predicted using language models and positional information. 
They clustered the remaining high-frequency relations into 117 relation predicates based on semantic similarities.
By comparing visual relation triplets in VrR-VG and the original Visual Genome metadata, those 117 predicates can be mapped back to 259 relation predicates in the original Visual Genome.

On the other hand, Zellers \textit{et al.} \cite{zellers2018neural} analyzed visual relations in Visual Genome and grouped them into four categories: 
\textbf{geometric} (e.g. above, behind, under),
\textbf{possessive} (e.g. has, part of, wearing),
\textbf{semantic} (e.g. carrying, eating, using), 
and \textbf{miscellaneous} (e.g. for, from, made of) 
(see more details in Sec. 3.1 and Table 1 of \cite{zellers2018neural}). 
The majority of the high-frequency relations in Visual Genome are geometric and possessive \cite{zellers2018neural}. 
Many of those relations can be easily predicted without visual information \cite{zellers2018neural, liang2019rethinking}. 
In contrast, semantic relations corresponding to activities are less frequent and hard to predict without visual information \cite{zellers2018neural}. In the aforementioned 259 relation predicates, 164 of them are identified by us as semantic relations:

\noindent
\textit{adorning, appearing in, approaching, are attached to, are sitting on, attached, attached to, attached to a, balancing on, biting, boarding, bordering, built into, catching, chasing, coming out of, crashing on, decorating, displayed on, displaying, draped over, drawn on, dressed in, drinking from, driving, driving down, driving on, eating from, entering, filled with, floating in, floating on, flying, flying a, flying above, flying in, flying over, flying through, going down, grabbing, grazing, grazing in, grazing on, gripping, hanging, hanging above, hanging from, hanging in, hanging off, hanging on, hanging on a, hanging out of, hanging over, hangs from, hangs on, hits, hitting, hung on, jumping, jumping on, laying, laying in, laying on, laying on a, leaning on, leaning over, licking, looking out, lying in, lying inside, lying next to, lying on, lying on top of, marking, mounted on, mounted to, moving, overlooking, painted, painted on, petting, playing, playing in, playing on, playing with, plays, pointing, printed on, reflected in, reflected on, reflecting, reflecting in, reflecting off, reflecting on, resting on, running in, running on, securing, selling, served on, serving, sewn on, sits in, sits on, sitting, sitting at, sitting behind, sitting in, sitting in a, sitting inside, sitting near, sitting next to, sitting on, sitting on a, sitting with, skiing, skiing down, skiing in, skiing on, sleeping on, sniffing, stacked on, standing inside, standing near, standing with, sticking out, sticking out of, stopped at, stuck in, stuck on, supporting, supports, surfing, surfing in, surfing on, swimming in, swinging, swinging a, swings, talking on, talking to, tied around, tied to, touching, waiting at, waiting on, walking, walking across, walking along, walking behind, walking down, walking in, walking near, walking next to, walking on, walking on a, walking through, walking to, walking up, walking with, working on, wrapped around, wrapped in, written on.}

COCO-test-VrR focuses on the semantic relations. We select 3,403 images from MSCOCO Karpathy 5K test split \cite{karpathy2015deep} where each image has at least one ground truth caption that contains at least one of the 164 semantic relation predicates. 
One ground truth caption that describes semantic relations is randomly sampled for each image.




\begin{figure*}[t!]
\begin{center}
\includegraphics[width=1.0\linewidth]{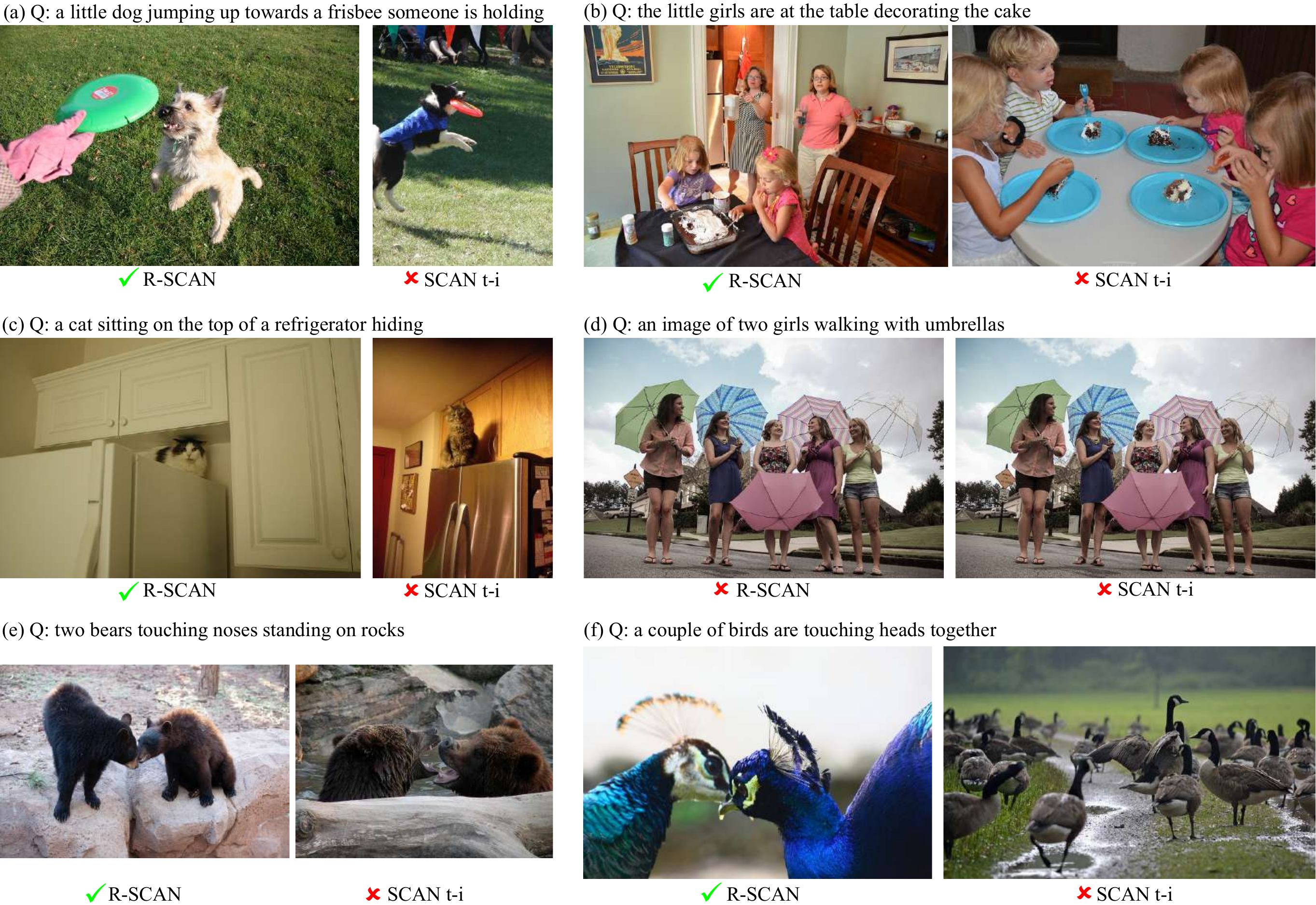}
\end{center}
   \caption{Additional qualitative examples of image retrieval given text queries on COCO-test-VrR using R-SCAN and SCAN t-i \cite{lee2018stacked} (both trained on MSCOCO). We show the top-1 ranked images for both models. Correctly retrieved images are marked with green check marks; incorrectly retrieved images are marked with red x. In (a) R-SCAN recognizes the frisbee is held by a person. In (b) R-SCAN recognizes the predicate ``decorating'' between the subject ``girls'' and the object ``cake''. In (c) R-SCAN is able to tell the cat is ``hiding'' on top of the refrigerator. (d) is an example where both R-SCAN and SCAN fail when image-text matching requires the ability of \textit{counting} (of the number of girls). This is relevant to the model's capability of visual reasoning \cite{johnson2017clevr} and remains to be addressed in future research works. In (e) R-SCAN identifies the activity ``touching noses'' and ``standing on rocks'', whereas SCAN t-i only gets the objects (bears, noses) and stuffs (rocks) right. (f) is an example where R-SCAN considers that two birds are ``touching'' each other's head, whereas SCAN t-i does not.}
\label{fig:t2i_vis}
\end{figure*}

\begin{figure*}[t!]
\begin{center}
\includegraphics[width=1.0\linewidth]{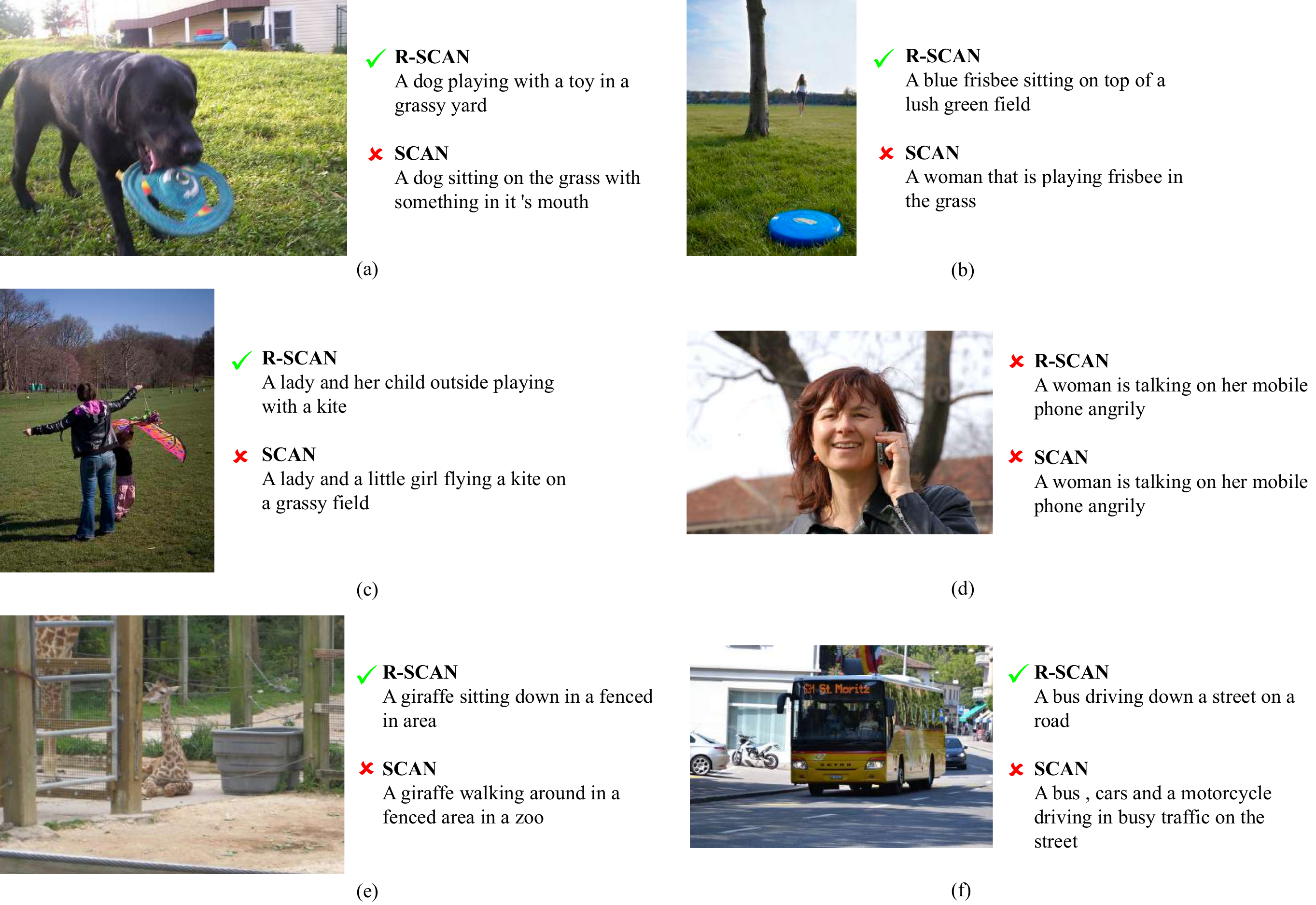}
\end{center}
   \caption{Additional qualitative examples of sentence retrieval given image queries on COCO-test-VrR using R-SCAN and SCAN t-i \cite{lee2018stacked} (both trained on MSCOCO). Image query is shown on the left of each example; top-1 ranked images are shown on the right. Correctly retrieved sentences are marked with green check marks; incorrectly retrieved sentences are marked with red x. In (a) the SCAN-retrieved caption incorrectly describes that the dog is ``sitting on'' grass. In (b) the SCAN result does not correctly describe the relation between ``woman'' and ``frisbee''. (c) is a tricky example where the lady and the little girl ``trying to fly'' a kite. (d) is an example where both R-SCAN and SCAN fail to retrieve the sentence that correctly describes emotion of the woman. This issue remains to be addressed in future research works. (e) is an example where the both R-SCAN and SCAN t-i recognize the object ``giraffe'' and stuff ``fenced area'' but SCAN fails to consider the semantic relation ``sitting down in''. In (f), although the SCAN result correctly matches the objects in image and text but the relations between the motorcycle and street or traffic are not correct. The motorcycle is not moving, but the SCAN-retrieved caption describes that it is ``driving in'' traffic. }
\label{fig:i2t_vis}
\end{figure*}

\begin{figure*}[t!]
\begin{center}
\includegraphics[width=1.0\linewidth]{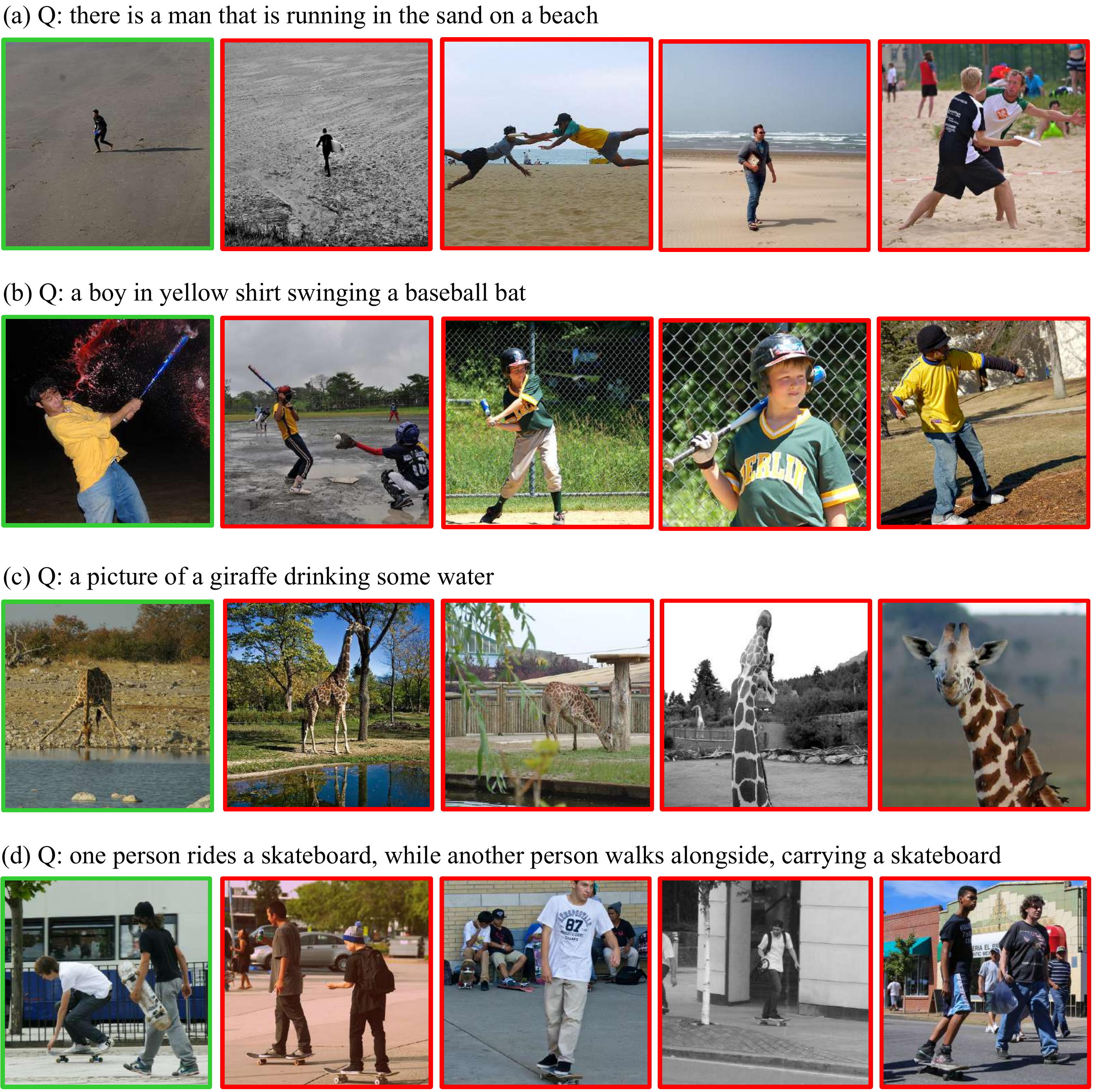}
\end{center}
   \caption{Qualitative examples of image retrieval given text queries on MSCOCO using R-SCAN. Each sentence description corresponds to one ground-truth image. For each sentence query, we show the top-5 ranked images, ranking from left to right. We outline the true matches in green and false matches in red. It can be observed that in these examples visual relations play important roles in ranking. For instance, in (c) the difference between the first and second place is the visual relation ``drinking'' as the subject ``giraffe'' and the object ``water'' present in both images.}
\label{fig:coco_t2i_topk_vis}
\end{figure*}

\begin{figure*}[t!]
\begin{center}
\includegraphics[width=1.0\linewidth]{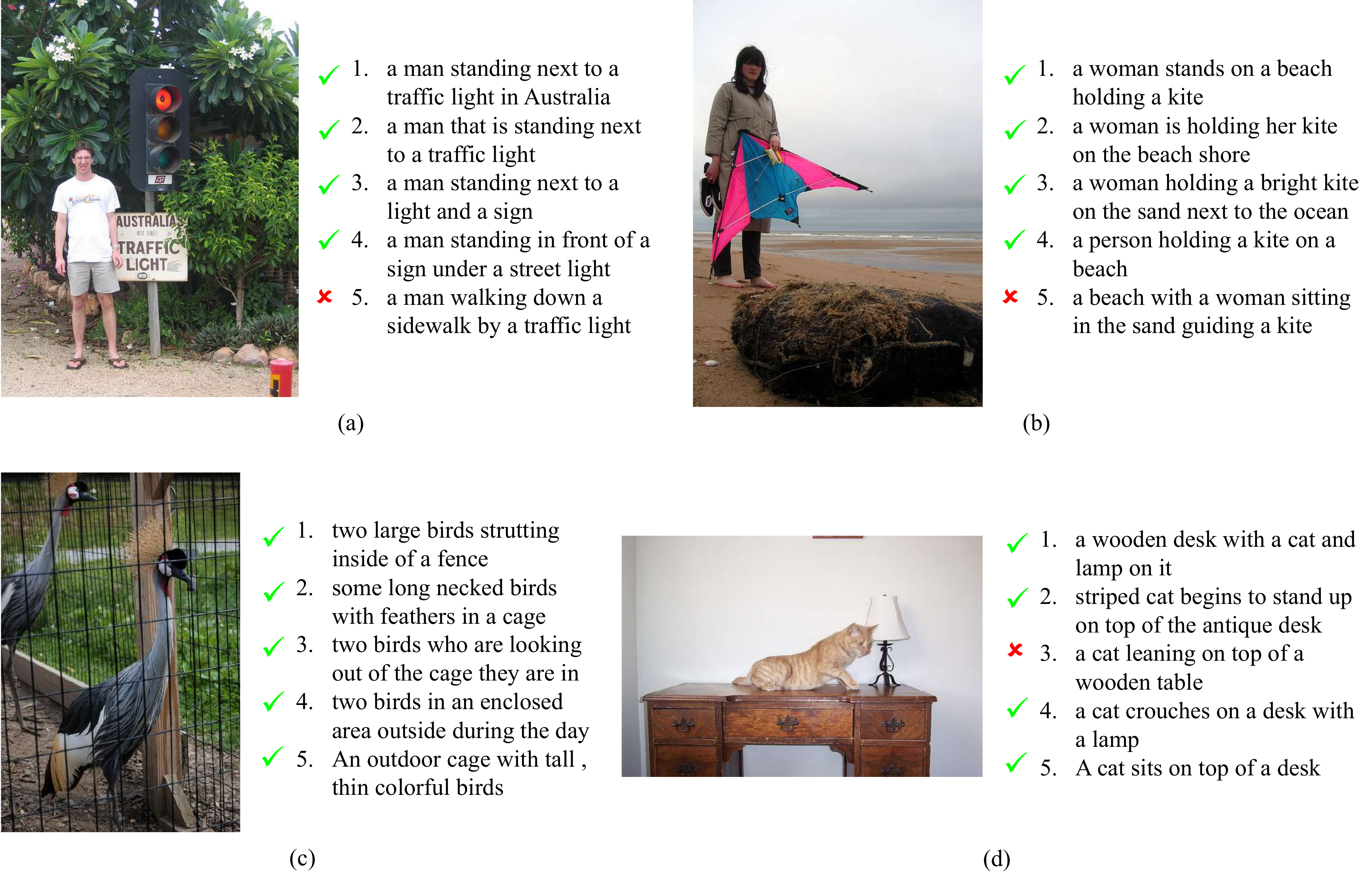}
\end{center}
   \caption{Qualitative examples of text retrieval given image queries on MSCOCO using R-SCAN. Correctly retrieved sentences are marked with green check marks; incorrectly retrieved sentences are marked with red x. Sentences are ordered by ranks in each example. These are examples where R-SCAN ranks most of the sentences that correctly describe visual relations higher than the incorrect descriptions. }
\label{fig:coco_i2t_topk_vis}
\end{figure*}

\begin{figure*}[t!]
\begin{center}
\includegraphics[width=1.0\linewidth]{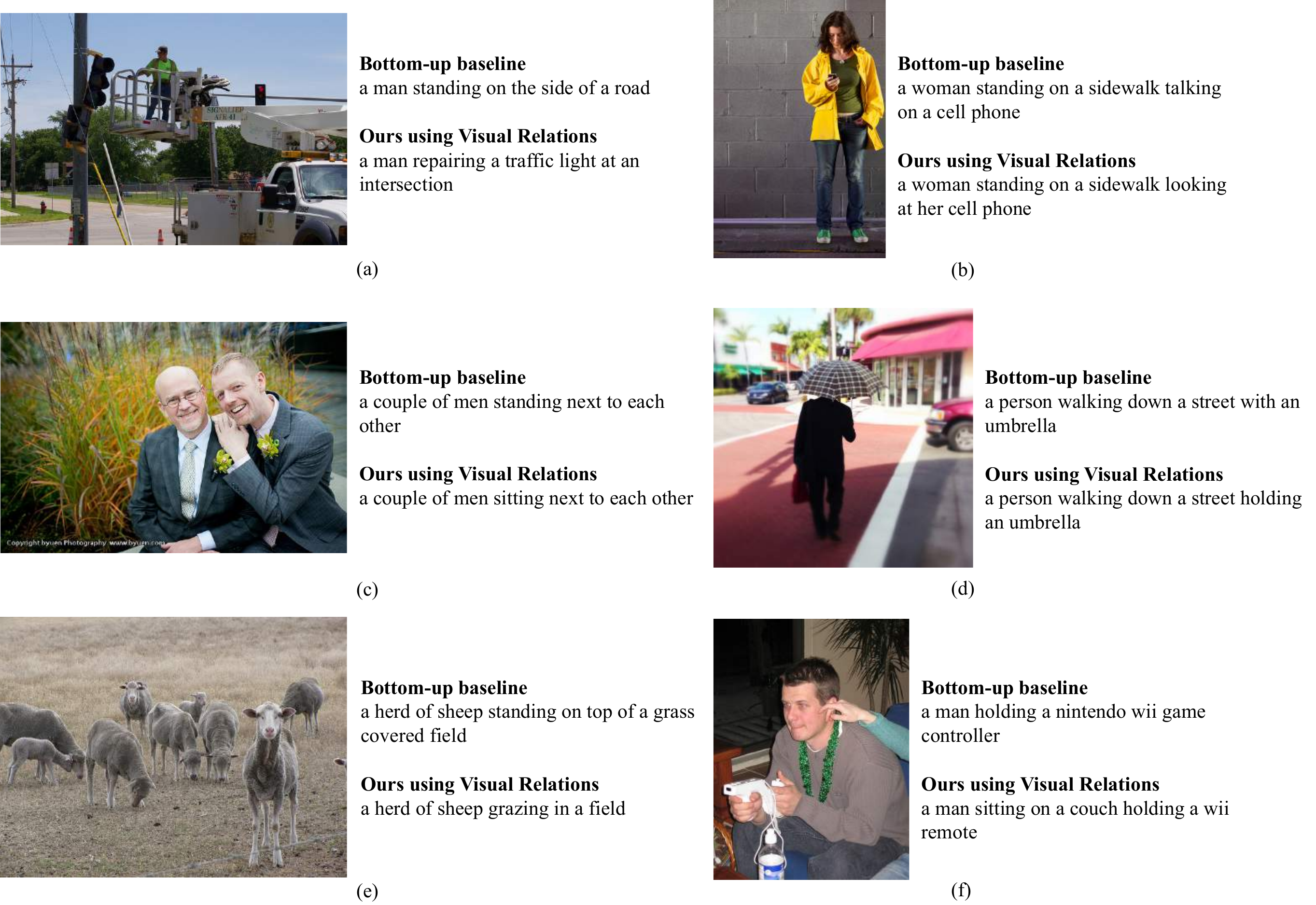}
\end{center}
   \caption{Qualitative examples of image captioning. We compare our captioning model using visual relations (introduced in Sec 4.4 of the main paper) with the bottom-up captioning model from \cite{anderson2017bottom} (both trained on MSCOCO). We show examples where our model predicts correct or more precise visual relations comparing to the bottom-up captioning model. For instance, in (b) our model is able to recognize the woman is ``looking at'' her cell phone instead of ``talking on'' it. }
\label{fig:caption_vis}
\end{figure*}

\section{Additional Qualitative Examples}
\label{more_r_coco_qual}

In this section, we present additional cross-modal retrieval examples to qualitatively demonstrate the effectiveness of incorporating visual relations for image-text matching. 
In Figure \ref{fig:t2i_vis}, we present the examples of image retrieval given text queries using R-SCAN and the baseline SCAN t-i AVG model \cite{lee2018stacked}. 
In Figure \ref{fig:i2t_vis}, we show the examples of text retrieval given image queries using R-SCAN and the baseline SCAN t-i AVG model. 
In Figure \ref{fig:coco_t2i_topk_vis}, we show the examples of top-5 ranked images given text queries on MSCOCO using R-SCAN. 
Similarly, in Figure \ref{fig:coco_i2t_topk_vis}, we show the examples of top-5 ranked sentences given image queries.

We also present qualitative examples of image captioning in Figure \ref{fig:caption_vis}. 
The top-down captioner baseline \cite{anderson2017bottom} is compared with our image captioning model using visual relation features (introduced in Sec 4.4 of the main paper).

\clearpage
\clearpage

{\small
\bibliographystyle{ieee}
\bibliography{egbib}
}

\end{document}